\documentclass{article} % For LaTeX2e
\usepackage{collas2022_conference,times}

\usepackage{amsmath,amsfonts,bm}

\def\eqref#1{equation~\ref{#1}}
\def\1{\bm{1}}

\DeclareMathAlphabet{\mathsfit}{\encodingdefault}{\sfdefault}{m}{sl}
\SetMathAlphabet{\mathsfit}{bold}{\encodingdefault}{\sfdefault}{bx}{n}

\newcommand{\E}{\mathbb{E}}

\DeclareMathOperator*{\argmax}{arg\,max}

\usepackage{url}
\usepackage[utf8]{inputenc} % allow utf-8 input
\usepackage[T1]{fontenc}    % use 8-bit T1 fonts
\usepackage{hyperref}       % hyperlinks
\usepackage{url}            % simple URL typesetting
\usepackage{booktabs}       % professional-quality tables
\usepackage{amsfonts}       % blackboard math symbols
\usepackage{nicefrac}       % compact symbols for 1/2, etc.
\usepackage{microtype}      % microtypography
\usepackage[pdftex]{graphicx}
\usepackage{xcolor}
\usepackage{subfigure}
\usepackage{wrapfig}
\usepackage{amsmath,bm}
\usepackage{amssymb}
\usepackage{mathtools}
\usepackage{tabularx}
\usepackage{lmodern}
\usepackage{hyperref}
\usepackage{wrapfig}
\usepackage{multicol}
\usepackage{enumitem}

\usepackage{algpseudocode} 
\usepackage[ruled,vlined]{algorithm2e}

\usepackage{amsmath}
\usepackage{cleveref} % needs to be after amsmath

\DeclarePairedDelimiterX{\infdivx}[2]{\big[}{\big]}{%
  #1\;\delimsize\|\;#2%
}
\usepackage[colorinlistoftodos]{todonotes}

\hypersetup{
    colorlinks=true,
    linkcolor=blue,
    filecolor=magenta,      
    urlcolor=blue,
    citecolor=purple,
    pdftitle={Overleaf Example},
    pdfpagemode=FullScreen,
    }

\title{Modelling non-reinforced preferences using selective attention} 

\author{Noor Sajid$^{1,2}$\thanks{Corresponding author: \texttt{noor.sajid.18@ucl.ac.uk}}, Panagiotis Tigas$^3$, Zafeirios Fountas$^2$, \\
$^1$WCHN, University College London, UK, $^2$Huawei Technologies, London, UK\\
$^3$OATML, Oxford University, UK\\
\And
Qinghai Guo$^4$, Alexey Zakharov$^2$, Lancelot Da Costa$^{1,5}$\\
$^4$Huawei Technologies, Shenzhen, China, $^5$Imperial College London, UK\\
}

\collasfinalcopy % Uncomment for camera-ready version, but NOT for submission.

\begin{document}
\maketitle

%, can exhibit these distinct behavioural modes---influenced by environment dynamics---to aptly trade-oﬀ between preference satisfaction and exploration. 

\begin{abstract}
How can artificial agents learn non-reinforced preferences to continuously adapt their behaviour to a changing environment? We decompose this question into two challenges: ($i$) encoding diverse memories and ($ii$) selectively attending to
these for preference formation. Our proposed \emph{no}n-\emph{re}inforced preference learning mechanism using selective attention, \textsc{Nore}, addresses both by leveraging the agent's world model to collect a diverse set of experiences which are interleaved with imagined roll-outs to encode memories. These memories are selectively attended to, using attention and gating blocks, to update agent's preferences. We validate \textsc{Nore} in a modified OpenAI Gym FrozenLake environment (without any external signal) with and without volatility under a fixed model of the environment---and compare its behaviour to \textsc{Pepper}, a Hebbian preference learning mechanism. We demonstrate that \textsc{Nore} provides a straightforward framework to induce exploratory preferences in the absence of external signals.

\end{abstract}

\section{Introduction}
Biological agents have the capacity to acquire preferences that inform their choices and lead to meaningful interactions with their environment. These preferences are a \textit{subjective assessment of what they would like to experience}---and can be continuously learnt, or modified, even in the absence of external feedback i.e, non-reinforced preferences~\citep{schonberg2020neural}. These are unlike reinforced preferences where preference for a stimulus may increase given some positive outcome associated with the stimulus i.e., reinforcement learning~\citep{sutton2018reinforcement}. Agents equipped with capacity to modify non-reinforced preferences are able to adjust their behaviour and adapt to changing environmental dynamics on the fly~\citep{schonberg2020neural,zajonc2001mere,izuma2013choice}---without needing to re-train, tune or optimise their world model. This motivates the current study: designing artificial agents equipped with a similar capacity to learn non-reinforced preferences that encourage adaptive behaviour.

Neuroscientific evidence reveals that non-reinforced preference changes can be driven by $i$) mere-exposure effects where frequently observed options are preferred~\citep{zajonc1968attitudinal,zajonc2001mere,grimes2007researching}, $ii$) attentional mechanisms where attending to an option can deem it more preferable~\citep{izuma2010neural,voigt2017endogenous,schonberg2020neural}, and $iii$) contextual effects where an option is preferred more compared to the alternative only in particular settings~\citep{izuma2013choice}. These preference shifts can be encoded using local plasticity rules e.g., Hebbian plasticity facilitates learning from mere-exposure effects~\citep{gerstner2002mathematical} or synaptic gating that can encourage enhancement of signal or suppression of noise~\citep{desimone1995neural,serences2014multi}. These speak to self-supervised learning mechanisms that support acquisition of non-reinforced preferences. 

\begin{wrapfigure}[23]{l}{0.7\textwidth}
    \centering
    \includegraphics[width=0.69\textwidth]{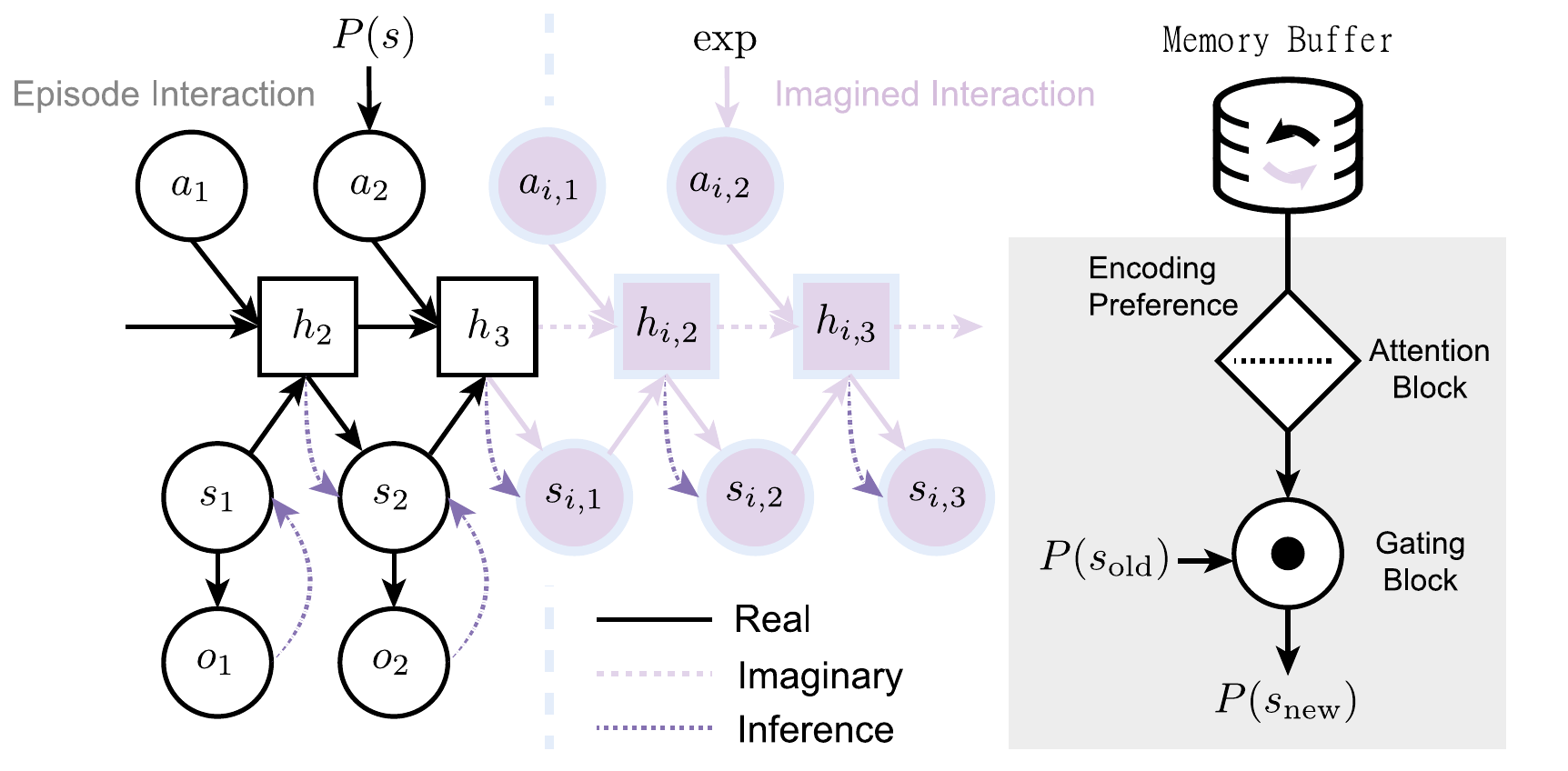}
    \caption{\textit{Preference learning mechanism.} \textsc{Nore} comprises two subsequent steps in each episode: $1$) environmental and imagined interaction, and $2$) accumulation of preferences once interaction ends using randomly shuffled memories. Both steps function in synergy: step $1$ influences preference learning and step $2$ influences the interactions in the subsequent episode.}
    \label{fig::preflearn}
\end{wrapfigure}

Here, we introduce \textsc{Nore}, a \textit{no}n-\textit{re}inforced preference learning mechanism that leverages synaptic gating to encode shifts in preferences in model-based agents. This allows the agent to learn distinct preferences that encourage adaptive behaviour at inference time. Briefly, \textsc{Nore} comprises a two-step procedure that occurs after the agent's world model has been optimised for the environment during training (see Figure~\ref{fig::preflearn}):
\begin{enumerate}[wide, labelwidth=!, labelindent=0.2pt]
    \item \textbf{Encoding memories.} For this, the agent has short episodes of direct exchange with the environment using the current non-reinforced preferences, and imagined interactions using an exploratory planner to find new novel states with high expected information gain (see \textsc{Plan2Explore}~\citep{sekar2020planning}). A history of latent state representations from a randomly selected subset ($30\%$) of \emph{real} environmental interactions are interleaved with \emph{imagined} interactions and retained---an abstraction of how memories are encoded using neural replay~\citep{breton2013memory}.
    \item \textbf{Encoding preferences using selective attention.} Once the agent's memory has been encoded, prior preferences are updated by optimising two blocks (attention and gating) via entropy maximisation~\citep{jaynes1957information}. The attention block constrains the agent's memories and mimics neural gain control via precision manipulation~\citep{rao2005bayesian,whiteley2008implicit,yu2004inference,meera2022reclaiming}. The gating block encodes the (prior) preference distribution by gating and storing relevant information in a time-dependent fashion~\citep{cho2014learning}. The outputs of this are used as prior preferences for the next environmental interaction. 
\end{enumerate}

\paragraph{Related work:}

The task of designing appropriate subjective objectives for the agent remains an open challenge in reinforcement learning. In the absence of any external stimuli, \textit{intrinsic motivation} has been proposed to guide the behaviour of an agent during its learning phase~\citep{shyam2019model, sekar2020planning, ball2020ready}. Intrinsic motivation has been formalised in a number of ways, using curiosity and surprise~\citep{Schmidhuber1991CuriousMC, Sun2011PlanningTB, Pathak2017CuriosityDrivenEB}, empowerment~\citep{Gregor2017VariationalIC}, information gain~\citep{Houthooft2016VIMEVI}, impact~\citep{Raileanu2020RIDERI}, or successor features~\citep{Zhang2019ScheduledID}. Importantly, the underlying objective of such agents is exploration for the purposes of learning the environment and the ways to exploit it (given extrinsic stimuli at test time). In contrast to the aforementioned exploration techniques, our work focuses on producing \textit{meaningful and adaptive} behaviour by learning subjective preferences over the states of the world at test time. 

A related work to \textsc{Nore} is \text{Pepper} \citep{sajid2021exploration} -- another preference learning mechanism that equips an agent with the ability to update preferences using Hebbian learning. This particular procedure restricts the agent's ability to appropriately filter (or gate) irrelevant information. Contrariwise, \textsc{Nore} encodes preferences reliant on the \textit{gating hypothesis}~\citep{schonberg2020neural} and learns to accentuate relevant information for a particular environment.

\section{Non-reinforced preference learning with selective attention (\textsc{Nore})}

\begin{wrapfigure}[22]{r}{0.3\textwidth}
\centering 
\includegraphics[width=0.27\textwidth]{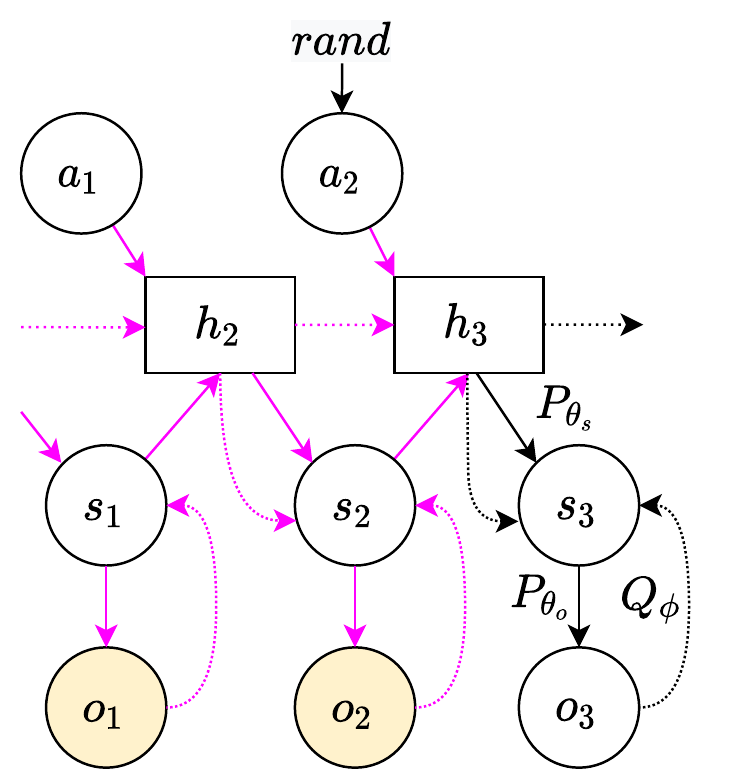}
\caption{\textit{Model architecture.} Circles and squares denote random and deterministic variables respectively. Coloured lines denote connections where learning occurs. Shaded circles represent outcomes that have already been observed by the agent.} \label{fig::model}
%\vspace{-25pt}
\end{wrapfigure} 
We aim to build an agent that can modify its preferences after learning about the environment without any reward signal nor supervision. This presents two challenges---encoding diverse memories and selectively attending to these for preference formation. We use the agent's world model and imagination to collect appropriate memories and optimise a Gated Recurrent Unit (GRU)~\citep{cho2014learning} to encode particular preferences. The preference learning procedure, \textsc{Nore}, is detailed in Algorithm~\ref{pseudocode}.
% \vspace{-15pt}
% \subsection{World model} 
\paragraph{World model.}
The agent's world model is instantiated as a Recurrent State-Space Model (RSSM)~\citep{hafner2019learning,hafner2019dream,hafner2020mastering,sajid2021exploration} and entails mapping a history of observations ($o_0, o_1,.., o_t)$ and actions ($a_0, a_1,.., a_t)$ to a sequence of deterministic states $h_t = f(h_{t-1}, s_{t-1}, a_{t-1})$ (Fig.~\ref{fig::model}). These deterministic states are used to calculate the latent prior and posterior states. These are used downstream for planning. Formally, the RSSM consists of the following:
($i$) GRU based deterministic recurrent model, $h_t=f_\theta(h_{t-1},s_{t-1}, a_{t-1})$; ($ii$) Latent state posterior, $Q_\phi(s_t|h_{t},o_t) \sim Cat$, and prior, $P(s) \sim Cat(D)$; ($iii$) Transition model, $P_\theta(s_t|h_t) \sim Cat$; ($iv$) Image predictor (or emission model), $Q_\phi(o_t|h_{t},s_t) \sim Bernoulli$. Here,  $Q_\phi(\cdot)$ denotes an approximate posterior distribution parameterised by $\phi$.

The world model is trained by optimising the evidence lower bound ($ELBO$)~\citep{hafner2019learning} or equivalently, the variational free energy \citep{friston2010free,fountas2020deep,sajid2021active, sajid2021exploration} using stochastic back-propagation with the Adam optimiser~\citep{kingma2014adam}. Here, the  training data is collected using trajectories generated under an exploratory policy. See Appendix \ref{appendix::efe} for $ELBO$ implementation.
\vspace{-.5em}

\paragraph{Planning objective for learning preferences.}
Following \cite{sajid2021exploration}, we substitute the planner with the expected free energy (G)~\citep{sajid2022active,barpGeometricMethodsSampling2022a} augmented with conjugate priors to allow for
preference learning over time:
\begin{align}\label{eq:G5a}
G(\pi,\tau) = &-\E\big[ \log P(o_{\tau}|s_{\tau},\pi) \big] 
+\E\big[ \log Q(s_{\tau}|\pi) - \log P(s|D) \big]+\E\big[ \log Q(\theta|s_{\tau},\pi) - \log P(\theta|s_{\tau},o_{\tau},\pi) \big],
\end{align}
%\vspace{-.8em}
%\begin{align}\label{eq:G5a}
%G(\pi,\tau) = &-\E_{\tilde{Q}}\big[ \log P(o_{\tau}|s_{\tau},\pi) \big] \\\label{eq:G5b}
%&+\E_{\tilde{Q}}\big[ \log Q(s_{\tau}|\pi) - \log P(s|D) \big]\\\label{eq:G5c}
%&+\E_{\tilde{Q}}\big[ \log Q(\theta|s_{\tau},\pi) - \log P(\theta|s_{\tau},o_{\tau},\pi) \big] ~ .
%\end{align}
where expectations are taken w.r.t. $Q_{\phi}(o_\tau, s_\tau, \theta | \pi) =Q(\theta|\pi)Q(s_{\tau}|,\pi)Q_\phi(o_{\tau}|s_{\tau},\pi)$ and the Dirichlet distribution was introduced as the conjugate prior over $Cat(D)$ (Appendix~\ref{appendix::dirichlet}). See Appendix \ref{appendix::g} for implementation of $G$. This planning objective bounds extrinsic and intrinsic value~\citep{da2020active,sajid2022active}. Therefore, in the absence of non-reinforced preferences, or whilst learning them, intrinsic motivation contextualises agent's interactions with the environment in a way that depends upon its posterior beliefs about latent environmental states~\citep{barto2013intrinsic,ryan2000intrinsic}. Here, actions are selected by sampling from the distribution $P(\pi) = \argmax(-G(\pi))$ ~\citep{barpGeometricMethodsSampling2022a}. 

%\begin{figure}[!t]
%    \centering
%    \includegraphics[width=0.80\linewidth]{figures/system_design.pdf}
%    \caption{\textit{Preference learning mechanism.} \textsc{Nore} comprises two subsequent steps in each episode: $1$) environmental and imagined interaction, \& $2$) accumulation of preferences once interaction ends using randomly shuffled memories. Both steps function in synergy: step $1$ influences preference learning and step $2$ influences the interactions in the subsequent episode.}
%    \label{fig::preflearn}
%\end{figure}

\paragraph{Selective attention during preference formation.}
Motivated by the neuroscientific view of attention as a gating mechanism~\citep{carrasco2011visual,meera2022reclaiming}, we introduce two blocks that gate preference encoding through signal accentuation and attenuation~\citep{desimone1995neural,serences2014multi}.
\begin{equation}
    \begin{array}{cc}
        \text{Attention: }  s \sim Q_\phi(s_{t-1}|h, \gamma) &  \text{Gating: } P_{\gamma}(s|h,D)
    \end{array}
\end{equation}
where, $\gamma$ denotes the precision term, $h$ the memory buffer and $D$ prior preferences. The attention block is a multi-layer perceptron (MLPs) and plays an analysis role to classical precision control mechanism. The gating block contains a deterministic (i.e., the recurrent state of the GRU) and stochastic component with diagonal covariance Gaussian distribution. 
These are trained during inference (i.e., preference learning) by maximising entropy over preferences using stochastic back-propagation with the Adam optimiser. The prior preferences used by the preference learning planner, $G$, are given by a softmax transformation. 

\section{Experiments}
Following \citep{sajid2021exploration}, we evaluated \textsc{Nore} on a variation of the OpenAI Gym FrozenLake environment~\citep{brockman2016openai} and compared its preference to \textsc{Pepper}~\citep{sajid2021exploration} at test time. Here, the agent is tasked with navigating a grid world comprised of frozen, hole and sub-goals, goal tiles, using four actions (left, right, down or up) without an external reward signal. Thus, although \textsc{Nore} and \textsc{Pepper} agents can differentiate between tile categories -- given their generative model -- no extrinsic signal is received from the environment. We used this environment to qualify how preferences evolved under the two preference learning mechanisms as a result of environmental volatility. For this, a volatile environment was simulated by modifying the FrozenLake grid every $K$ steps and initialising the agent in a different location at the start of each episode. This provides an appropriate test-bed to assess how much volatility was necessary to induce shifts in learnt prior preferences. The world model weights were frozen during these experiments. Therefore, any behavioural differences are a direct consequence of \textsc{Pepper} that induces differences in the planning objective, G. See Appendix \ref{appendix::traingenmodel} for architecture and training details for each environment.
\begin{figure}[!h]
    \centering
    \includegraphics[width=0.85\linewidth]{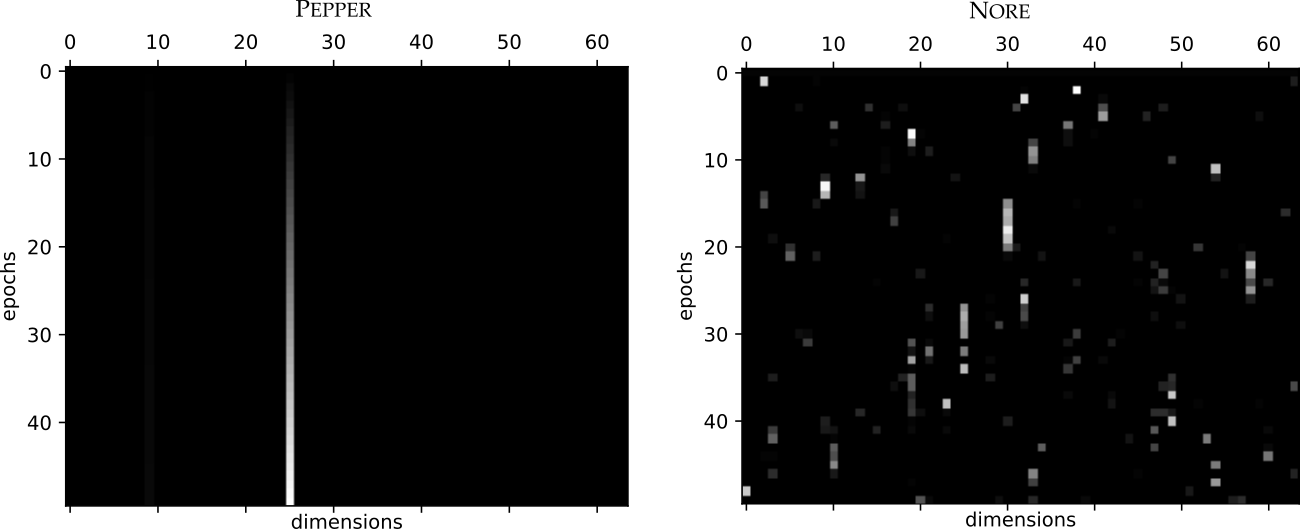}
    \caption{Sample learnt state preferences for a single \textsc{Nore} and \textsc{Pepper} agent in a static environment. Here, $64$ state categories are presented on the x-axis and episodes on the y-axis. The scale goes from white (high Dirichlet concentration) to black (low Dirichlet concentration) and grey indicates gradations between these.}
    \label{fig::1}
\end{figure}
\paragraph{Encoded non-reinforced preferences.} First, we compared the preferences encoded by \textsc{Nore} and \textsc{Pepper} agents when the environment was static i.e., no changes to the grid configuration. Figure~\ref{fig::1} plots the difference between the two preference learning rules. For the Hebbian learning rule (\textsc{Pepper}), preferences are reinforced given particular environmental exposure i.e., the more the agent experiences something, the more it is preferred. Conversely, for the preference learning via synaptic gating (\textsc{Nore}) preferences shift across epochs -- with new preferences being encoding until the last epoch.   

\begin{figure}[!h]
    \centering
    \includegraphics[width=0.95\linewidth]{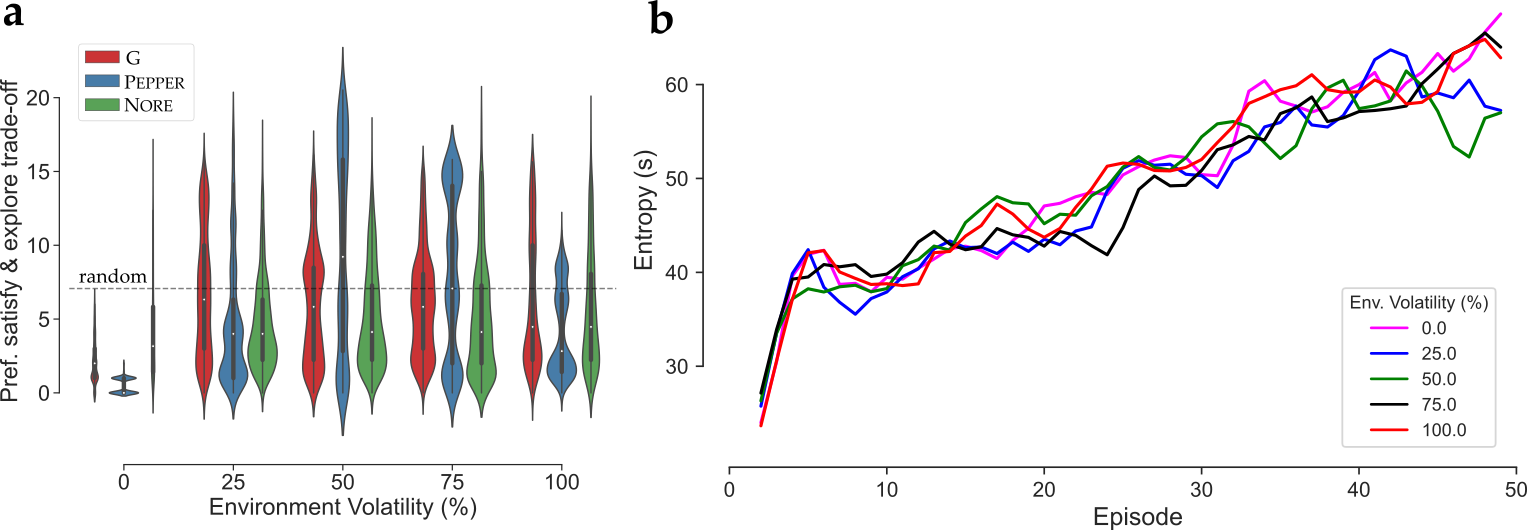}
    \caption{\textbf{A:} The violin plot scores the exploration and preference satisfaction for varying degrees of environmental volatility. The x-axis denotes environmental volatility from none ($0\%$ i.e., constant map), to change every $25\%$ , $50\%$, $75\%$ and every step. The y-axis denotes the Hausdorff distance (Appendix~\ref{appendix::hausdorff}). Red denotes the agent optimising the standard expected free energy (G); blue is \textsc{Pepper}; and green is \textsc{Nore}. \textbf{B:} The line plot depicts the entropy of $P(s)$ over time for varying levels of environmental volatility. The x-axis counts episodes, and the y-axis entropy in natural units.}% The pink line is for $0\%$, blue for $25\%$, green for $50\%$, black for $75\%$ and red for $100\%$ volatility in the environment.}
    \label{fig:tradeoff}
\end{figure}

\paragraph{\textsc{Nore} behaviour.} We evaluated the agent's behaviour in Figure \ref{fig:tradeoff} using Hausdorff distance (Appendix~\ref{appendix::hausdorff})~\citep{blumberg1920hausdorff}. With this metric we observed increased exploration by the \textsc{Nore} agent at $50\%, 75\%$ volatility in the environment. Similar to \textsc{Pepper}, the agents pursued long paths from the initial location (see Fig.~\ref{fig::frozenlake} for sample trajectories). Compared to \textsc{Pepper}, \textsc{Nore} agents did not exhibit bi-modal preferences at $100\%$ volatility in the environment.

\section{Concluding remarks}

We introduced \textsc{Nore}, a new non-reinforced preference learning mechanism using selective attention. \textsc{Nore} leverages the agent's world model and imagined roll-out to encode appropriate memories and selectively attends to them for preference learning.  An agent equipped with this non-reinforced preference learning mechanism has the capacity to continuously learn and modify its subjective assessment of what is preferred; which can induce exploratory behaviours. Practically, further downstream these agents might accomplish tasks specified via simple and
sparse rewards more quickly, or may acquire broadly useful
skills that could be adapted to specific task objectives.

%\paragraph{Connections to plasticity rules}

%Such a mechanism would rely on areas causally related to the deployment of overt attention. Causal roles have been demonstrated for the macaque lateral intraparietal (LIP) cortex [59,60], the frontal eye fields (FEF) [61], and the superior colliculus (SC) [
 
%We need evaluation data to allow for appropriate representation; 
%Cheaper
%Offline tuning / separation 

\bibliography{collas2022_conference}
\bibliographystyle{collas2022_conference}
\newpage
\appendix

\section{Pseudo-code for learning preferences}
%%% Pseudo code for the preference learning
\begin{algorithm}[H]
\label{pseudocode}  
 \caption{\textsc{Nore}}
\SetEndCharOfAlgoLine{}
 \SetKwComment{Comment}{// }{}
 \SetKwInOut{Input}{Input}
 \Input{\\\hspace{-3.6em}\small
    \begin{tabular}[t]{l @{\hspace{.25em}} l}%
        $h_t := f_\theta(h_{t-1},s_{t-1}, a_{t-1})$ & Recurrent model\\
        $Q_\phi(s_t|h_t,o_t)$ & Posterior model\\
        $Q_\phi(s_t|h_t)$ & Prior model\\
        $P_\theta(o_t|h_t,s_t)$ & Observation model
    \end{tabular}%
  }
 \textbf{Initialise} \;
 uniform Dirichlet prior over $P(s)$ \tcc*[r]{prior preference being learnt}
 learning rate $\alpha$; memory trace $\beta$; \; 
 \For{\textup{each episode} e}{
         reset environment and collect initial observations ($o_0$)\;
         \For{ \textup{each time step} t \tcc*[r]{environment interaction}}{  
          compute $s_{po} \sim Q_\phi(s_t|h_t,o_t)$ \;
          compute G using $P(s)$, observed and predicted posteriors \;
          $a_t \leftarrow \argmax(-G(\pi)$) \;
          \textbf{execute} $\sim  a_t$ and receive $o$\;
          $o_{t+1} \leftarrow o$\;
         }
         \For{\textup{each imagined step \tcc*[r]{imagination}}}{{
          compute $s^i_{po} \sim Q_\phi(s_t|h_t,o_t)$ \;
          $a_t \sim \mathcal{A}$ \;
          compute $s^i_{pr}, s^i_{po}$ using $h_t \& a_t$\;
         }}
        $s^m_{po} \leftarrow \text{comb}(\text{filter}(s_{po}), s^i_{po})$  \tcc*[r]{re-coding memories}
        \For{\textup{each memory \tcc*[r]{encoding preferences}}}{{
        $s^a_{po} \leftarrow \text{MLP}(s^m_{po})$  \tcc*[r]{attention block}
        $w \leftarrow \text{GRU}(s^a_{po}, w)$ \tcc*[r]{gating block}
        $P(s) \leftarrow \beta * P(s) + \alpha * w$
         }}
        maximise $-\mathbb{E}_{P(s)}\log P(s)$ 
    }
\end{algorithm}

\section{Implementation}\label{appendix::efe}
Our generative model was implemented exactly as in Pepper~\citep{sajid2021exploration} using Dreamer V2's public implementation~\citep{hafner2020mastering}\footnote{\url{https://github.com/danijar/dreamerv2}}. Specifically, Dreamer's generative model training loop was used, alongside a model predictive control (MPC) planner. Therefore, the actor learning part of Dreamer was not incorporated, and the generative model was trained using Plan2Explore~\citep{sekar2020planning}. Like Plan2Explore, an ensemble of image encoders were learnt and the ``disagreement'' of the encoders was used as an intrinsic reward during training. This guides the agent to explore areas of the map that have high novelty and potentially high information gain, when acquiring a generative (i.e., world) model. See appendix $\mathcal{A}$ in~\citep{sajid2021exploration} for further details regarding the planner.  

Upon training completion, we froze the generative model's learnt weights and only allowed learning of prior preferences. These preferences were updated after each episode as described in Algorithm~\ref{pseudocode}.

\subsection{Evidence lower bound}
The generative model was optimised using the ELBO formulation introduced in \citep{hafner2019dream}:

\begin{align}\label{e:elbo_appendix}
     \mathcal{L}(\theta)=\sum\limits_{t=1}^T &
     \underbrace{-\mathbb{E}\left[ \log P_\theta(o_t\mid s_t,\pi) \right]}_{\mbox{reconstruction}}
     + \underbrace{\mathbb{E}\left[ D_{KL}(Q_\phi(s_t\mid o_t, s_{t-1}, \pi)) \parallel P_\theta(s_t\mid s_{t-1},\pi) \right]}_{\mbox{dynamics}} ~ ,%\nonumber ~ .
     \vspace{-1em}
\end{align}
where both expectations are w.r.t. $Q_\phi(s_t\mid o_{\leq t}, a_{\leq t})$.

%\begin{align}\label{e:elbo_appendix}
%     \mathcal{L}(\theta)=\sum\limits_{t=1}^T \big[&
%     \underbrace{\underset{Q_\phi(s_t\mid o_{\leq t}, a_{\leq t})}{-\mathbb{E}\left[ \log P_\theta(o_t\mid s_t,\pi) \right]}}_{\mbox{reconstruction}}
%     + \underbrace{\underset{Q_\phi( s_t\mid o_{\leq t}, a_{\leq t} )\qquad \qquad\qquad\qquad\qquad}{\mathbb{E}\left[ D_{KL}(Q_\phi(s_t\mid o_t, s_{t-1}, \pi)) \parallel P_\theta(s_t\mid s_{t-1},\pi) \right]}}_{\mbox{dynamics}} \big]\nonumber ~ .
%     \vspace{-1em}
%\end{align}

\subsection{Expected free energy}\label{appendix::g}
We implemented G using the parameterisations introduced in \citep{sajid2021exploration}. Explicitly, for each term in~\eqref{eq:G5a}:

\begin{itemize}

    \item \textbf{Term 1} was computed as the entropy of the observation model $P(o_\tau|s_\tau,\pi)$. Happily, the factorisation of the observation model -- as independent Gaussian distributions -- allowed us to calculate the entropy term in closed form.
    
    \item \textbf{Term 2} was computed as the difference between $\log Q(s_\tau|\pi)$ and $\log P(s_\tau|D)$, where $\log Q(s_\tau|\pi)$ was approximated using a single sample from the prior model $Q(s_\tau|\theta,\pi)$. Again, the dependency on $\pi$ was substituted by $h_t$.
    
    \item \textbf{Term 3} was computed by rearranging the expression to $H(o_\tau|s_\tau,\theta,\pi)-H(o_\tau|s_\tau,\pi)$ \citep{fountas2020deep,sajid2021exploration}. This translates to $I(o_\tau;\theta|s_\tau,\pi)$, and can be approximated using Deep Ensembles~\citep{lakshminarayanan2016simple,sekar2020planning} and calculating their variance $\text{Var}_\theta[\mathbb{E}Q(o_\tau|s_\tau,\theta,\pi)]$. Here, each ensemble component can be seen as a sample from the posterior $Q(\theta|s_\tau,\pi)$. Our experiments showed that using 5 components was sufficient.
    
\end{itemize}

\section{Hausdorff distance}\label{appendix::hausdorff}

This metric calculates the maximum distance of the agents' position in a particular trajectory to the nearest position taken in another trajectory. Thus high Hausdorff distance denotes increased exploration, since trajectories observed across episodes differ from one another. Contrariwise, a low distance entails prior preference satisfaction as agents repeat trajectories across episodes.

\section{Dirichlet distribution}\label{appendix::dirichlet}
Here, we used the Dirichlet distribution as the conjugate prior over $Cat(D)$ with dimension $n\times n $ defined as:
\vspace{-.4em}
\begin{align}
    P(D^{i}|d^{i}) = Dir(d^{i}) \Rightarrow \begin{cases}
      \mathbb{E}_{P(D^{i}|d^{i})} \big[D^i_{ij}\big]=\frac{d^i_{ij}}{\sum_k d^i_{kj}} \\
       \mathbb{E}_{P(D^{i}|d^{i})} \big[log(D^i_{ij})\big]=\digamma(d^i_{ij})-\digamma(\sum_k d^i_{kj})
    \end{cases}  
    \vspace{-.2em}
\end{align}
where, $\digamma$ is the digamma function, $d \sim \mathbb{R}^+$.

%% Should add the derivation for this! 
%\begin{align*}
 %   MLP([x,y])&= [\omega \odot x, y] \iff MLP([x,y]) = \begin{bmatrix}
  %  \omega &0\\
  %  0& Id \end{bmatrix} \begin{bmatrix}x\\y \end{bmatrix}\\
  %  \text{input to GRU} &= MLP([x,y]) \\ % [Dense > Norm > Activation]
  %  \text{output to GRU} &= softmax(deter) \equiv P_{\gamma}(s|h; PD)
%\end{align*}

\section{Experiments}\label{appendix::traingenmodel}
We simulated the agent in five distinct situations ranging from a non-volatile, static environment to a highly volatile one i.e., a different FrozenLake\footnote{\url{https://github.com/openai/gym/} (MIT license)} map every step. For all episodes in the static setting, the agent was initialised at a fixed location with no changes to the FrozenLake map throughout that particular episode. Conversely, agents operating in the volatile setting were initialised at a different location each time. Moreover, the FrozenLake map was also changed every $N$ steps -- given the desired volatility level (Table \ref{table:training_param}). These experiments were deliberately kept simple to gain an understanding of how non-reinforced preferences could be learnt using selective attention and how they differed from hebbian preference learning rules. Future work should investigate how \textsc{Nore} agents behave in more complex, open-ended environments.  

\begin{table}[h!]
\caption{Training parameters}
\centering
\begin{tabular}{l|l|l}
\textbf{Parameter}&\textsc{Nore}&\textsc{Pepper}\\
\midrule
 Planning Horizon & $15$ steps & $15$ steps \\
 Episode Length & $100$ steps  & $100$ steps \\
 Reset Every & $10$ steps & $10$ steps \\
 No. Episodes & $50$ episodes & $50$ episodes \\
 No. State Categories & $64$ categories & $64$ categories \\
 No. State Dimensions & $50$ dimensions & $50$ dimensions \\
\end{tabular}\label{table:training_param}
\end{table}

\begin{table}[h!]
\caption{Preference learning parameters }
\centering
\begin{tabular}{l|l|l}
\textbf{Parameter}&\textsc{Nore}&\textsc{Pepper}\\
\midrule
 Planning Horizon & $15$ steps & $15$ steps\\
 Episode Length & $100$ steps & $100$ steps\\
 No. Episodes & $50$ episodes & $50$ episodes\\
 Reset Map Every & $1,25,50,75,100$ steps  & $1,25,50,75,100$ steps \\
\end{tabular}
\end{table}

\subsection{\textsc{Nore} behaviour}\label{appendix::dirichlet}

\begin{figure}[!h]
    \centering
    \includegraphics[width=0.59\textwidth]{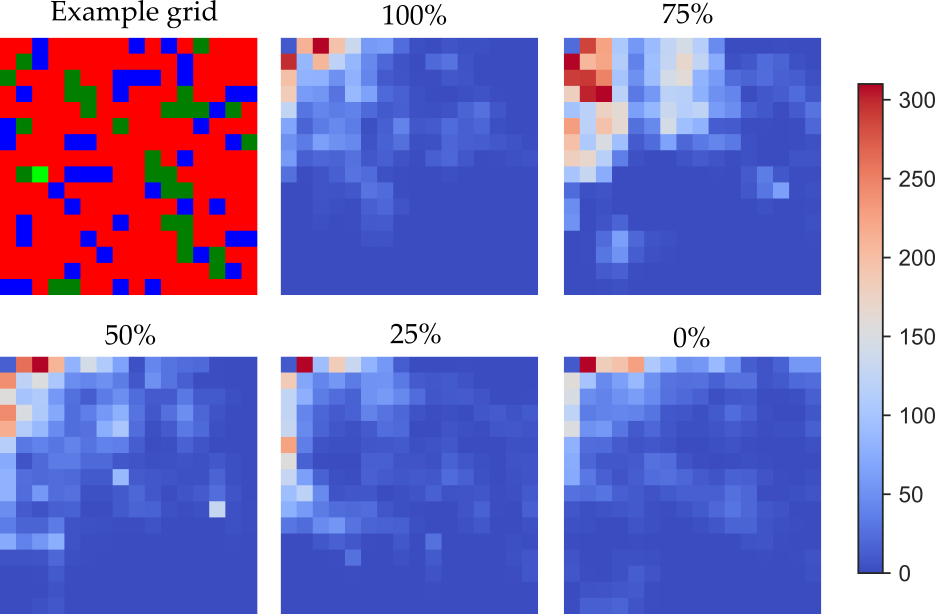}
    \caption{\textit{FrozenLake and environmental trajectories} }\label{fig::frozenlake}
\end{figure}

\end{document}